\documentclass[conference]{IEEEtran}
\IEEEoverridecommandlockouts

\usepackage{cite}
\usepackage{amsmath,amssymb,amsfonts}
\usepackage{graphicx}
\usepackage{textcomp}
\usepackage{xcolor}

\usepackage{algorithm}
\usepackage{algpseudocode}
\usepackage{amsmath, amsfonts}
\usepackage{bm}
\usepackage{booktabs}
\usepackage{multirow}
\usepackage{hyperref}
\usepackage{tcolorbox}
\usepackage{subcaption}
\usepackage{url}

\DeclareMathOperator*{\argmax}{\arg\!\max}

\usepackage[capitalize]{cleveref}
\crefname{table}{Table}{Tabs.}
\crefname{figure}{Fig.}{Figs.}
\Crefname{section}{Section}{Sections}
\Crefname{table}{Table}{Tables}
\Crefname{assumption}{Assumption}{Assumptions}
\crefname{algorithm}{Algorithm}{Algorithms}

\def\BibTeX{{\rm B\kern-.05em{\sc i\kern-.025em b}\kern-.08em
    T\kern-.1667em\lower.7ex\hbox{E}\kern-.125emX}}
\begin{document}

\title{Fair In-Context Learning via Latent Concept Variables}

\author{\IEEEauthorblockN{Karuna Bhaila\textsuperscript{1}, Minh-Hao Van\textsuperscript{1}, Kennedy Edemacu\textsuperscript{1}, Chen Zhao\textsuperscript{2}, Feng Chen\textsuperscript{3}, and Xintao Wu\textsuperscript{1}}
\IEEEauthorblockA{
\textsuperscript{1}\textit{University of Arkansas}, Fayetteville, AR, USA \\
\textsuperscript{2}\textit{Baylor University}, Waco, TX, USA \\
\textsuperscript{3}\textit{University of Texas at Dallas}, Richardson, TX, USA \\
kbhaila@uark.edu, haovan@uark.edu, kedemacu@uark.edu, chen\_zhao@baylor.edu, \\feng.chen@utdallas.edu, xintaowu@uark.edu
}}

\maketitle

\maketitle
\begin{abstract}
The emerging in-context learning (ICL) ability of large language models (LLMs) has prompted their use for predictive tasks in various domains with different data types, including tabular data, facilitated by serialization methods. However, with increasing applications in high-stakes domains, it has been shown that LLMs can inherit social bias and discrimination from their pre-training data. In this work, we investigate inherent bias in LLMs during in-context learning with tabular data. We focus on an optimal demonstration selection approach that utilizes latent concept variables for resource-efficient task adaptation. We design data augmentation strategies that reduce the correlation between predictive outcomes and sensitive variables, helping promote fairness during latent concept learning. We utilize the learned concept to select demonstrations and obtain fair predictions. The latent concept variables are learned using a smaller internal LLM and generalized to larger external LLMs. We empirically verify that the fair latent variable approach improves fairness results on tabular datasets compared to multiple heuristic demonstration selection methods. Code and data are available at \url{https://github.com/karuna-bhaila/fairicl}.
\end{abstract}


\section{Introduction}
LLMs have demonstrated immense capabilities in performing various natural language processing (NLP) tasks. A factor contributing to widespread LLM usage is their in-context learning~\cite{brown2020language} ability, which allows adaptation to downstream tasks without costly training or fine-tuning. With a few demonstration examples, ICL equips LLMs with the ability to infer task-specific context and perform inference with impressive utility. Recent research has also explored the applicability of LLMs on tabular data through serialization methods that facilitate ICL by transforming the data into natural language formats~\cite{hegselmann2023tabllm}. With the increasing integration of LLM inference in domains such as healthcare~\cite{wu2023medical}, finance~\cite{li2023large}, and the legal sector~\cite{sun2023short} with various data formats, it has become crucial to scrutinize their use from a trustworthiness perspective.

LLMs have been shown to exhibit discriminatory behavior in their outputs due to stereotypes and prejudices inherent in pre-training data~\cite{abid2021large, basta2019evaluating}. When used for decisive tasks, LLMs may mirror social inequalities and biases from the real world, leading to harmful consequences. Furthermore, in ICL settings with tabular data classification, recent research has empirically verified the presence of bias in LLM outputs. For instance,~\cite{liu2023ConfrontingLW} investigated unfairness in ICL with tabular data and evaluated mitigation by flipping labels of in-context demonstration examples, resulting in bias reduction but with significant trade-offs in model utility.~\cite{li2024chatgpt} similarly implemented multiple heuristic methods for demonstration selection based on sensitive attributes and label distribution in the demonstrations.~\cite{hu2024strategic} discovered that increasing the representation of minority groups and underrepresented labels in demonstrations helps to improve fairness with utility costs and developed a strategy that uses clustering to extract representative samples for selecting demonstrations based on performance on a validation set. 
In this work, we similarly explore optimal demonstration selection for ICL to promote fairness in LLM predictions, but utilize the latent concept variable mechanism ~\cite{wang2024large} to achieve fair in-context learning. ~\cite{wang2024large} formulated ICL via a Bayesian perspective and theorized that inference with a finite number of demonstrations selected using latent concept approximates the optimal Bayes predictor. The latent concept is learned from an observed set of task-specific training data with a small LLM and used to obtain demonstrations that can be generalized to larger LLMs for improving performance. 

Motivated by the influence of latent concepts on model performance, we formulate a fair demonstration selection approach for in-context learning, dubbed as \textbf{FairICL}. As the latent concepts are learned from an observed set of task-specific examples, we conjecture that the training data distribution may affect the quality of the learned latent concepts and ultimately the model predictions from both accuracy and fairness perspectives. Therefore, in FairICL, we incorporate an effective data augmentation technique that promotes decorrelation between the sensitive attributes and the outcome variables by randomizing the relationship between them. This augmentation allows us to obtain a fairer representation of the task-specific data used to learn the fair latent concept variables while preserving relevant information among non-sensitive attributes and the label. We then utilize the learned concepts to select demonstrations from the observed training examples such that the probability of observing the learned latent variable is maximized when conditioned on the corresponding example. The selected demonstrations are used to perform in-context learning with external LLMs larger than the one used for learning. This framework can support private businesses or organizations to obtain fair LLM predictions on their local data without having to train/fine-tune large models with fairness objectives.
We empirically validate FairICL on real-world tabular datasets known to represent social biases and demonstrate that FairICL can effectively achieve fairness goals while maintaining predictive utility. We compare the performance of FairICL with multiple heuristic approaches and conduct a comprehensive analysis of the influence of different hyperparameters. Our empirical results show that FairICL can generalize demonstration selection to external LLMs and outperform baseline methods.

\section{Related Work}
\label{sec:rel_work}
As LLM integration into decision-making systems grows, it becomes essential for them to be evaluated from a fairness perspective. Multiple works have highlighted discriminatory behavior in LLM outputs originating from societal biases contained in pre-training data~\cite{abid2021large, wang2023decoding} or under-representation of minority population~\cite{gallegos2023survey}. For instance,~\cite{huang2021implicit} analyzed implicit gender-based stereotypes in LLM outputs via commonsense inference.~\cite{wang2023decoding} evaluated the influence of normal and adversarial prompts on bias in GPT models.~\cite{abid2021large} demonstrated unfairness in LLM outputs with respect to religious groups. 

Following these works, the study of LLM fairness has also extended to tabular data inference with pre-trained language models~\cite{li2023fairness, liu2023ConfrontingLW, chhikara2024few, hu2024strategic, atwood2024inducing}. These works focus on LLM inference with in-context learning and formulate ways to select demonstration examples while promoting fairness or ensuring representation for minority groups.~\cite{li2023fairness} evaluated multiple heuristic methods of selecting demonstrations and a guardrail technique instructing LLM to be fair in its decision-making.~\cite{liu2023ConfrontingLW} implemented label-flipping for demonstration examples and achieved fair predictions but with a large utility loss.~\cite{chhikara2024few} evaluated LLM's familiarity with commonly known fairness notions and utilized a similarity-based demonstration selection approach.~\cite{hu2024strategic} aimed to increase minority group representation in demonstrations and selected demonstrations based on corresponding validation set performance.~\cite{atwood2024inducing} explored remediation techniques for fairness and compared prompt-based techniques with in-processing and post-processing methods. 
Similar to some earlier works, we aim to address bias in LLM predictions in tabular data by exploring optimal demonstration selection for in-context learning to promote fairness in LLM predictions. However, we utilize the latent concept variable model and present a framework to learn fair representations of the latent concept, which will guide the demonstration selection process instead of relying on heuristic methods such as manual prompting or label flipping.

\begin{figure*}[!t]
    \centering
    \includegraphics[width=0.95\textwidth]{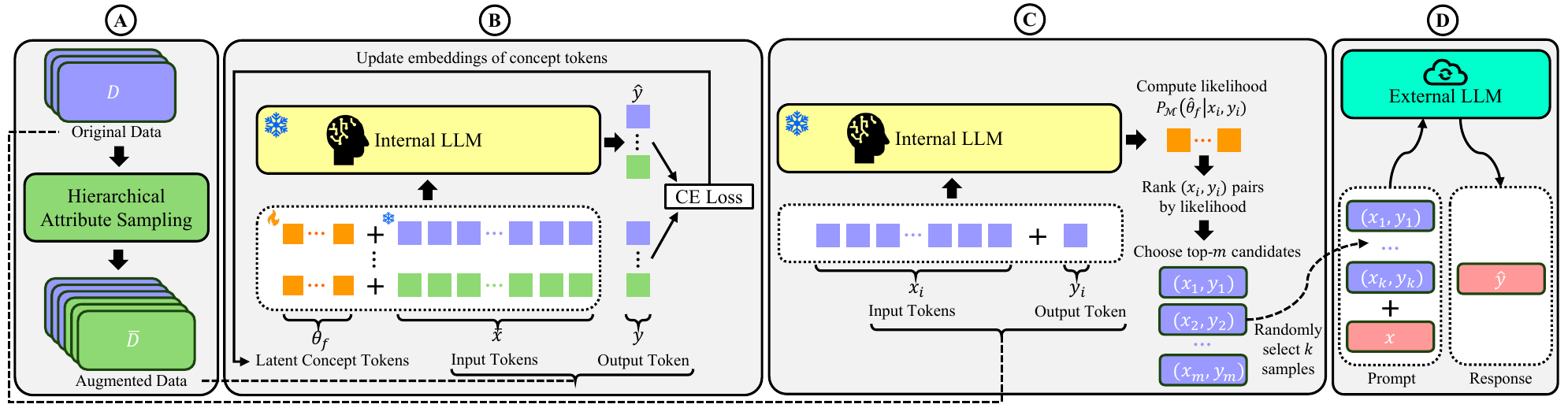}
    \caption{Overview of FairICL including steps from \textbf{A} to \textbf{D}, \textbf{A}: A hierarchical attribute sampling approach is used to create augmented training data $\bar{D}$; \textbf{B}: Samples in $\bar{D}$ are utilized to learn latent concept tokens with an internal LLM; \textbf{C}: A corresponding likelihood score is computed for each sample in $D$, and all samples are then ranked to choose $k$ demonstrations from top-$m$ candidates; \textbf{D}: Selected demonstrations and test input $x$ are used to prompt an external LLM to get prediction $\hat{y}$.}
    \label{fig:fairicl}
\end{figure*}

\section{Preliminaries}

\subsection{In-Context Learning}
The in-context learning~\cite{brown2020language} ability of LLMs has prompted multiple works that investigate how LLMs learn from demonstration examples for certain tasks without being explicitly trained. 
Let us denote a pre-trained LLM as $\mathcal{M}$ with parameters $\mathbf{W}$. Let $D = \{(x_i, y_i)\}_{i=1}^n$ denote a tabular dataset observed for an arbitrary task where $x_i \in \mathcal{X}$ represents attributes of the $i$-th instance and $y_i \in \mathcal{Y}$ its corresponding outcome. Assume $a_i \in \mathcal{A}$ denotes its sensitive attribute. For ICL, the LLM is provided with $k$ examples from $D$ as demonstrations to guide the model in structuring its response for a test example $x$. Conditioned on a task description $inst$, a set of demonstrations $\{(x_1,y_1), \cdots, (x_k,y_k)\}$, and a test query $x$, the prediction output $\hat{y}$ from $\mathcal{M}$ can be shown as
\begin{equation}\label{eq:icl}
\hat{y} \leftarrow \mathcal{M}(inst, \underbrace{g(x_1, y_1),\cdots,g(x_k, y_k)}_\text{demonstration examples}, g(x)),
\end{equation}
where $g(x_k, y_k)$ denotes a prompt function (e.g., a template) that transforms the $k$-th demonstration into natural language text. To simplify, we omit the task description and prompt function thereafter and represent the output probability as
\begin{align}
    P_\mathcal{M}(y|(x_1,y_1),\cdots,(x_k,y_k),x;\mathbf{W}).
\end{align}
ICL performance has been found to be significantly influenced by demonstration examples and their ordering~\cite{liu2021makes, rubin2021learning, su2022selective, lu2021fantastically, ma2024fairness}. Consequently, recent works explore effective demonstration selection based on similarity to query input~\cite{liu2021makes, rubin2021learning, su2022selective}, entropy of predicted labels~\cite{lu2021fantastically}, and low predictive bias~\cite{ma2024fairness}.

\subsection{Latent Concept Learning}
An essential research question in in-context learning is effective demonstration selection to enable optimal downstream performance. Towards this objective,~\cite{xie2021explanation} put forth an interpretation of ICL based on latent concept variables, concluding that pre-trained models learn latent concepts during next-token prediction training and infer a shared latent concept among demonstrations used for inference. They show that under assumptions of a hidden Markovian data generation process, discrete latent concept tokens, and an approximately infinite number of demonstrations, in-context learning is an optimal predictor. Similarly,~\cite{wang2024large} studied latent concepts in LLMs but under a more general assumption of continuous latent concepts and that the data generation process is governed by an underlying causal mechanism given as $X \rightarrow Y \leftarrow \theta$ or $Y \rightarrow X \leftarrow \theta$ where $X$ represents the input, $Y$ the output, and $\theta$ the latent concept variable. Then, in-context learning can become an optimal predictor with a finite number of demonstrations chosen using the latent concept variable $\theta$. To find the optimal value of $\theta$ when considering the $X \rightarrow Y \leftarrow \theta$ direction,~\cite{wang2024large} formulated a latent concept variable framework for learning task-specific concept tokens that capture sufficient information for next-token prediction by minimizing a loss jointly conditioned on $X$ and the learned $\hat{\theta}$ as
\begin{align}\label{eq:theta}
    l(x,y;\hat{\theta}) = -\operatorname{log}P_\mathcal{M}(y|\hat{\theta}, x),
\end{align}
where $\hat{\theta}$ represents the learned latent concept variable, $x$ denotes an input token sequence and $y$ the discrete target variable. In practice, $\hat{\theta}$ is optimized by adding new tokens to $\mathcal{M}$'s vocabulary with corresponding embedding vectors, which we refer to as $\mathbf{W}_\theta$. During training, $\mathbf{W}_\theta$ is updated using the loss defined above. The learned $\hat{\theta}$ is then used to select $k$ most suitable demonstrations based on the likelihood of observing the concept tokens when conditioned on the demonstration pairs formulated as $P_\mathcal{M} (\hat{\theta} | (x_i, y_i), \hdots, (x_k, y_k))$. Assuming independence among the sampled demonstrations, the top-ranked examples are obtained based on latent concept likelihood for individual examples,
\begin{align}\label{eq:likelihood}
    \argmax_{(x_i, y_i) \in D} P_\mathcal{M} (\hat{\theta} | x_i, y_i).
\end{align}
The selected demonstrations are used to perform in-context learning and are further generalizable for inference with LLMs larger than the ones used to learn $\hat{\theta}$.

\section{Fair Latent Concept Learning}\label{sec:fairicl}
LLMs have been shown to replicate bias and prejudice likely present in their pre-training corpus. Providing LLMs with biased examples as demonstrations during ICL may further corroborate the prediction bias, potentially leading to discriminatory outcomes in classification tasks. However, filtering pre-training data and re-training/fine-tuning LLMs on unbiased data is often practically infeasible due to resource constraints. Moreover, removing discrimination from pre-training data may not entirely address the unfairness resulting from biased demonstrations during inference. Here, we focus on the demonstration selection process, which can guide LLM predictions by providing task-specific contextual information. Researchers have empirically shown that varying demonstrations can affect the bias and fairness outcomes of LLMs~\cite{hu2024strategic, ma2024fairness}. Furthermore, the proportion of samples from minority and majority groups in demonstrations affects the trade-off between fairness and performance metrics~\cite{hu2024strategic}. 

\subsection{Problem Setup}
In the latent concept variable model, demonstrations are selected based on the likelihood of observing latent concept variable $\hat{\theta}$~\cite{wang2024large}. Generally, the concept variables capture format and task information and can help improve in-context learning performance. However, the quality of the learned latent concept variables highly depends on the observed data $D$. We hypothesize that using a biased dataset $D$ to learn the latent concept can lead to selecting demonstrations that favor the majority group. For instance, consider a dataset containing a comparatively higher number of positive/advantaged class instances for the majority group, reflecting real societal bias. The latent concept variables may associate the positive outcome with the majority class, as this biased prediction can lead to better prediction accuracy owing to imbalanced label distributions. Consequently, demonstrations selected using these latent concept variables can reinforce the bias in the dataset. In the following, we propose FairICL, a fair latent concept learning framework with data augmentation to mitigate unfairness in ICL predictive outcomes arising from demonstration selection. An overview of the method is presented in~\cref{fig:fairicl}.

\begin{algorithm}[htb]
\caption{Fair Latent Concept Learning and Demonstration Selection}
\label{alg:fairicl}
\begin{algorithmic}[1]
    \renewcommand{\algorithmicrequire}{\textbf{Input:}}
    \Require Training dataset $D$, generated dataset $\tilde{D}$, test query $x$, LLM $\mathcal{M}$, number of tokens $c$, learning rate $\lambda$, training epochs $T$, number of demonstrations for training $q$, number of demonstration candidates $m$, number of demonstrations for inference $k$
    \renewcommand{\algorithmicrequire}{\textbf{Output:}}
    \Require $k$ demonstrations for test query $x$
    \Statex \texttt{\textcolor{blue}{/* Constructing Augmented Data */}} 
    \For{$(x_i,y_i) \in \{D \cup \tilde{D}\}$}
        \State Sample $(x_1,y_1),\cdots,(x_q,y_q)$ from $D$
        \State $\bar{x}_i = (x_1,y_1,\cdots,x_q,y_q,x_i)$
        \State Add $(\bar{x}_i, y_i)$ to $\bar{D}$
    \EndFor
    \Statex \texttt{\textcolor{blue}{/* Learning Fair Latent Concept */}} 
    \State Add $c$ new tokens to $\mathcal{M}$'s vocabulary representing $\theta_f$
    \State Freeze $\mathcal{M}$'s pre-trained parameters and initialize $\mathbf{W}_{\theta_f}$
    \For{$t = 1 \dots T$}
        \For{$(\bar{x}_i,y_i) \in \bar{D}$}
            \State $\hat{y}_i = P_\mathcal{M}(y_i|\hat{\theta}_f,\overline{x}_i)$
        \EndFor
        \State $\mathcal{L} = -\sum\limits_{i = 1}^{n+\tilde{n}}\operatorname{log}P_\mathcal{M}(y_i|\hat{\theta}_f,\overline{x}_i)$
        \State $\widehat{\mathbf{W}}_{\theta_f} \leftarrow \widehat{\mathbf{W}}_{\theta_f} - \lambda\frac{\partial\mathcal{L}}{\partial \mathbf{W}_{\theta_f}}$
    \EndFor
    \Statex \texttt{\textcolor{blue}{/* Selecting demonstrations */}}    
    \For{$(x_i,y_i) \in D$}
    \State Calculate likelihood as $P_\mathcal{M}(\hat{\theta}_f|x_i,y_i)$
    \EndFor
    \State Sort $D$ by likelihood
    \State Select top-$m$ $(x,y)$ pairs as demonstration candidates
    \State Randomly choose $k$ demonstrations from candidate set
    \State \textbf{Return} Demonstrations for test sample $x$
\end{algorithmic}
\end{algorithm}

\subsection{Constructing Augmented Training Data}\label{sec:augmented}
To ensure fair predictive outcomes, we consider the correlation between the sensitive attribute $a$ and the outcome variable $y$ in the dataset $D$ used to learn the latent concept variable $\theta$. We conjecture that learning latent concept variables from an unbiased dataset can prevent $\hat{\theta}$ from incorporating bias into the task-specific contextual information that improves ICL performance. To this end, we design and implement a data pre-processing strategy on $D$ aimed at decorrelating the sensitive attribute and the label. Assuming we obtain a dataset $\tilde{D}$ that preserves task-relevant information from $D$ and not the biased correlation between $a$ and $y$, we then construct an augmented training dataset $\bar{D}$ from both $D$ and $\tilde{D}$ to promote fairness while learning task-specific contextual information in a fair representation of latent concepts $\hat{\theta}_f$. Note that our focus is on LLM classification with ICL on tabular data, which is the most commonly used data representation in fairness literature.

To simulate decorrelation, we generate synthetic examples sampled from the distribution of $D$ but with any influence of $a$ on $y$ removed. Naively, we can achieve this decorrelation by sampling each attribute in $x$ and the label $y$ independently. However, this could result in highly noisy data samples that obfuscate useful information about the relationship between non-sensitive attributes and the label, thus negatively affecting the prediction accuracy. Here, we propose to generate samples based on an attribute hierarchy-based sampling process that simultaneously promotes sensitive attribute and label decorrelation and preserves task-relevant information. For hierarchical attribute sampling, we define an order for non-sensitive attributes and construct a synthetic sample based on this order. First, we randomly sample a label from a uniform distribution and obtain a subset of $D$ conditioned on the sampled label value. We then uniformly sample the first non-sensitive attribute in the ordered list from the values occurring in the subset. We further constrain the subset to include only the sampled value of the first non-sensitive attribute, and sample the second non-sensitive attribute uniformly, and so on. To populate the sensitive attribute value, we randomly sample it from a uniform distribution independent of the label and any non-sensitive attributes. Furthermore, if $D$ contains any proxy-sensitive attributes that may allude to the sensitive attribute, we condition its sampling on the sensitive attribute value to promote complete decorrelation. In this manner, we generate $\tilde{D} = \{(x_i,y_i)\}_{i=1}^{\tilde{n}}$ as an unbiased representation of $D$. This process is outlined in lines 1-5 in~\cref{alg:fairicl}.

We construct our augmented training dataset comprised of $n+\tilde{n}$ instances, each containing $q$ demonstration examples from $D$ and one query sample from either $D$ or $\tilde{D}$ to facilitate ICL. Formally each instance takes the form $\left<(x_1,y_1), \cdots, (x_q, y_q), x, y\right>$ which we denote as $(\overline{x},y)$ thereafter. We also denote this formatted dataset containing augmented samples as $\bar{D} = \{\bar{x}_i, y_i\}_{i=1}^{n+\tilde{n}}$. 
The following discusses how we learn the fair latent concept variables from $\bar{D}$.

\subsection{Learning Fair Latent Concept Variable}\label{sec:learning}
We learn the latent concept variables by implementing prompt tuning to optimize a set of new token embeddings that is prepended to each training input token sequence~\cite{wang2024large}. More importantly, we utilize the augmented dataset $\bar{D}$ to construct input sequences for learning $\theta_f$ to promote improvements in fairness and utility simultaneously. Directly optimizing $\theta_f$ as a sequence of words is inefficient due to the discrete nature of text space. Typically, large language models (LLMs) process inputs as sequences of tokens, which are subsequently transformed into embeddings. Therefore, we optimize the fair latent concept in the LLM $\mathcal{M}$'s embedding space, where $\theta_f$ is represented as a sequence of $c$ learnable tokens, each associated with an embedding vector. We denote the subset of weights in $\mathbf{W}$ corresponding to $\theta_f$ as $\mathbf{W}_{\theta_f}$. 
During training, we prepend $\hat{\theta}_f$ to the input sequences and learn $\mathbf{W}_{\theta_f}$ by minimizing the negative log-likelihood objective as follows
\begin{align}
    \mathcal{L} = -\sum\limits_{i = 1}^{n+\tilde{n}}\operatorname{log}P_\mathcal{M}(y_i|\hat{\theta}_f,\overline{x}_i). 
\end{align}
During gradient optimization, parameters $\mathbf{W}_{\theta_f}$ corresponding to $\hat{\theta}_f$ are updated, and all other parameters are frozen. The ultimate goal of learning fair latent concept variables is to derive the optimal $\widehat{\mathbf{W}}_{\theta_f}$ using the task-specific data $D$ to improve performance and the generated data $\tilde{D}$ to promote fairness simultaneously. In~\cref{alg:fairicl}, the latent concept learning process is shown in lines 6-14.

\begin{table*}[htb]
    \centering
    \caption{Dataset statistics}
    \begin{tabular}{c|c|cc|cc|cc}
    \toprule
    \multirow{2}{*}{Dataset} & \multirow{2}{*}{Label} & \multicolumn{2}{c|}{Train $(D)$} & \multicolumn{2}{c|}{Augmented $(\tilde{D})$} & \multicolumn{2}{c}{Test ($D^{t}$)} \\
    \cmidrule{3-8}
     & & Priv. & Unpriv. & Priv. & Unpriv. & Priv. & Unpriv. \\
     \midrule
     \multirow{2}{*}{Adult} & Positive &  6363 & 1105 & 7574 & 7468 & 250 & 250 \\ 
                            & Negative & 13897 & 8635 & 7534 & 7424 & 250 & 250 \\ 
     \cmidrule{1-8}
     \multirow{2}{*}{COMPAS} & Positive &  654 & 1333 & 1061 & 1089 & 125 & 125 \\ 
            & Negative & 1036 & 1199 & 1042 & 1030 & 125 & 125 \\ 
     \cmidrule{1-8}
     \multirow{2}{*}{LawSchool} & Positive & 11281 & 527 & 3263 & 3314 & 125 & 125 \\ 
               & Negative &   965 & 311 & 3309 & 3198 & 125 & 125 \\ 
     \bottomrule
    \end{tabular}
    \label{tab:dataset}
\end{table*}

\begin{figure*}[htb]
\centering
    \begin{subfigure}[h]{0.75\linewidth}
    \includegraphics[width=\linewidth]{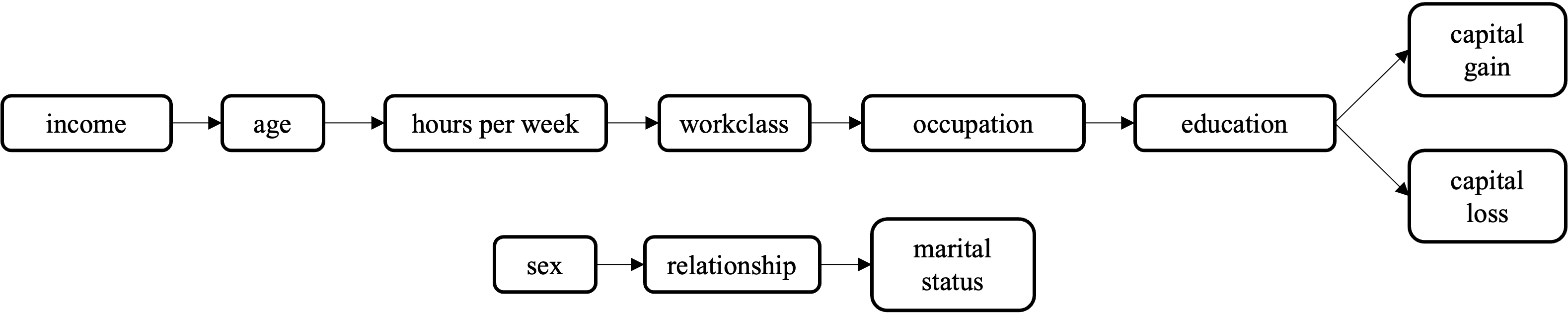}
    \caption{Adult dataset}
    \label{fig:adult_hierarchy}
    \end{subfigure}

    \vspace{1.0em}

    \begin{subfigure}[h]{0.75\linewidth}
    \includegraphics[width=\linewidth]{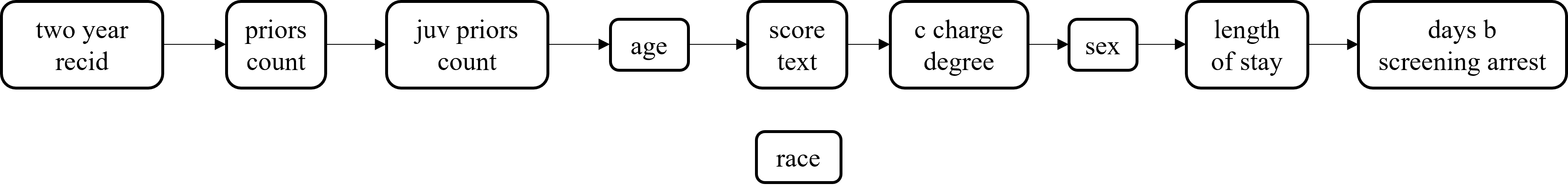}
    \caption{COMPAS dataset}
    \label{fig:compas_hierarchy}
    \end{subfigure}

    \vspace{1.0em}

    \begin{subfigure}[h]{0.75\linewidth}
    \includegraphics[width=\linewidth]{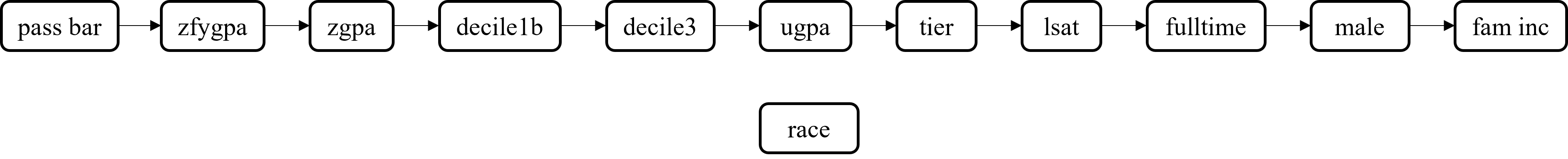}
    \caption{LawSchool dataset}
    \label{fig:law_hierarchy}
    \end{subfigure}
\caption{Hierarchical order of attributes for augmented data generation}
\label{fig:hierarchy}
\end{figure*}

\subsection{Demonstration Example Selection with $\theta_f$ Likelihood}\label{sec:selection}
The learned fair latent concept $\hat{\theta}_f$ is then used to select top-ranking examples from $D$, which will be provided as context to a larger external LLM during inference via ICL. This demonstration selection follows the rationale that training examples that maximize the likelihood of predicting the trained task-specific latent concept variables are optimal demonstrations for the corresponding task objective~\cite{wang2024large}.
For each training example $(x_i,y_i) \in D$, the likelihood of $\hat{\theta}_f$ is expressed using the probability distribution shown as
\begin{equation}\label{eq:dem_selection}
    P_\mathcal{M}(\hat{\theta}_f|x_i,y_i).
\end{equation}
In our implementation, we obtain this likelihood as the probability of observing the trained $\hat{\theta}_f$ when postfixed to a sample $(x_i,y_i)$. In~\cref{alg:fairicl}, these steps are outlined in lines 15-17. 
Subsequently, training examples are sorted based on their computed likelihood values. We then select the top $m$ examples that maximize the likelihood of $\hat{\theta}_f$ and form the demonstration candidate set. We subsample this candidate set to allow each test query to be paired with varying demonstrations during testing. Finally, we randomly select $k$ demonstration examples from the candidate set for each test instance, combining these to construct the final prompt for inference with an external LLM. 
The augmented dataset generation, latent concept learning, and demonstration selection procedures are summarized in~\cref{alg:fairicl}.

\section{Experimental Evaluation}

\begin{figure*}[htb]
    \centering
    \begin{tcolorbox}[colback=gray!10,left=2pt, right=2pt, top=2pt, bottom=2pt, fontupper=\small, fontlower=\small]
    \footnotesize
    \#\#\# Instruction: Based on the profile description of an individual recorded in the 1994 U.S. census, answer the question about their income.\\\\
    \#\#\# Profile: This person is a 38 years old female. She has attended a professional school and works in the private sector. Her occupation is in professional specialty. She works 50 hours per week. She had a capital gain of 0 and a capital loss of 0 last year. She has never been married. She is not related to the other person in her household. \\\\
    \#\#\# Question: Does this person have an income of more than 50,000 U.S. dollars? \\
     \#\#\# Answer: Yes \\
    $\vdots$ \\
    \#\#\# Profile: This person is a 28 years old male. He has an academic associate's degree and works in local government. His occupation is in protective services. He works 40 hours per week. He had a capital gain of 0 and a capital loss of 0 last year. He is married to a civilian. He is the husband of the other person in his household. \\\\
    \#\#\# Question: Does this person have an income of more than 50,000 U.S. dollars? \\
    \#\#\# Answer: 
    \end{tcolorbox}
    \caption{Serialization and prompt format for tabular Adult Income dataset.}
    \label{fig:serialization}
\end{figure*}

\subsection{Datasets}
We evaluate the effectiveness of fair demonstration selection with FairICL using three benchmark fair machine learning datasets: Adult Income dataset~\cite{barry1996adult}, COMPAS~\cite{compas}, and LawSchool~\cite{quy2022survey}. The respective binary prediction tasks are to predict whether an individual has an annual income greater than 50,000 US dollars based on demographic and economic attributes, predict the risk of recidivism based on a defendant's screening survey responses, and predict whether a student passes the bar exam based on their admission records. 

As discussed in~\cref{sec:augmented}, we generate an augmented dataset to enable fair latent concept learning. For Adult dataset, we follow previous work~\cite{liu2023ConfrontingLW} and use a subset of 10 attributes. We consider \textit{sex} as the sensitive attribute and distinguish \textit{relationship} and \textit{marital status} as proxy-sensitive attributes, as some instances of \textit{relationship} attribute contain gender-specific vocabulary and attribute \textit{marital status} may depend on values of \textit{relationship}. For COMPAS and LawSchool datasets, we assign \textit{race} as the sensitive variable. To generate augmented samples, we specify a hierarchical order for the non-sensitive attributes and a separate order for the sensitive and proxy-sensitive attributes based on the analysis in~\cite{quy2022survey}; the respective orders are shown in~\cref{fig:hierarchy}. As COMPAS and LawSchool datasets do not contain proxy-sensitive attributes, we sample only the sensitive attribute independently. Using this attribute sampling technique, we generate $\tilde{n} = n$ number of unique augmented data samples and construct our training dataset $\bar{D}$ by combining $D$ and $\tilde{D}$. For the test dataset, we randomly sample 1000 instances with equal representation for majority and minority groups for each experimental run. The statistics for the training, augmented, and test datasets are included in~\cref{tab:dataset}. For each record in the dataset, we perform serialization on the tabular dataset similar to~\cite{hegselmann2023tabllm, carey2024dp}, i.e., we convert each row in the dataset to a natural language format to facilitate LLM prompting. An example serialization template and in-context learning format is shown in~\cref{fig:serialization}.

\subsection{Fairness Metrics}\label{sec:fairness}
Here, we briefly describe two fairness notions used to determine LLM's bias w.r.t the majority and minority groups represented by sensitive attribute $a$ when predicting a binary outcome variable $y$.

\paragraph{Statistical Parity}
Statistical parity~\cite{dwork2012awareness} requires the predictions to be independent of the sensitive attribute and can be evaluated as
\begin{align}\label{eq:sp}
    \Delta \text{SP} = P(\hat{y}|s = 0) - P(\hat{y}|s = 1)
\end{align}

\paragraph{Equal Opportunity}
Equal opportunity~\cite{hardt2016equality} requires that for members of majority and minority groups, the probability of being assigned a positive outcome is the same. We evaluate equal opportunity using group-based TPRs as
\begin{align}\label{eg:eo}
    \Delta \text{EO} &= P(\hat{y}=1|y=1,s=0) - P(\hat{y}=1|y=1,s=1)
\end{align}
For datasets where the negative outcome is the favorable one, we evaluate $\Delta$EO as the difference in group-based TNRs.

\subsection{Baselines}\label{sec:baselines}
We compare FairICL against several baselines that implement different demonstration selection approaches. \textit{Random} refers to standard in-context learning where $k$ demonstrations are randomly sampled for each test instance~\cite{brown2020language}. \textit{Balanced} implements in-context learning with equal representation for each sensitive attribute and class label combination in the demonstrations~\cite{li2023fairness}. \textit{Instruction} is used to evaluate an LLM for fair and unbiased decisions based on manual prompting-based guidance with a balanced demonstration set~\cite{li2023fairness, atwood2024inducing}. \textit{Removal} omits the sensitive attribute from the demonstrations of Balanced~\cite{li2023fairness}. As we serialize tabular data, we further replace gendered pronouns with gender-neutral ones in the training data. \textit{Counterfactual} is another heuristic technique and constructs demonstrations using $k/2$ examples from the majority (minority) group and the remaining examples by flipping the sensitive attribute of the previously sampled examples~\cite{li2023fairness}. \textit{LatentConcept} is the approach from~\cite{wang2024large} where the latent concept variables are learned using the training dataset and then used to select top-$k$ demonstrations.

\begin{table*}[htb]
    \centering
    \caption{Performance and fairness metrics of FairICL on the Adult dataset compared with baselines on LLaMA-2-7B and LLaMA-2-13B as external LLMs and LLaMA-2-7B as the internal LLM for latent concept learning; bold denotes best performance among fairness-aware methods and underline denotes best performance among all models}
    \begin{tabular}{c|c|c|c|c|c}
\toprule
External LLM & Method & Acc(\%)$\uparrow$ & F1(\%)$\uparrow$ & $|\Delta$SP$|\downarrow$ & $|\Delta$EO$|\downarrow$ \\
\midrule
\multirow{7}{*}{LLaMA-2-13B} & Random~\cite{brown2020language} & 76.00\textsubscript{1.19} & 75.75\textsubscript{1.44} & 0.14\textsubscript{0.04} & 0.11\textsubscript{0.08} \\
                             & LatentConcept~\cite{wang2024large} & \underline{77.48}\textsubscript{0.70} & 77.22\textsubscript{0.74} & 0.16\textsubscript{0.02} & 0.12\textsubscript{0.01} \\
\cmidrule{2-6}
                             & Balanced~\cite{li2023fairness} & 74.58\textsubscript{1.38} & 71.92\textsubscript{2.19} & 0.13\textsubscript{0.05} & 0.10\textsubscript{0.07} \\
                             & Counterfactual~\cite{li2023fairness} & 68.18\textsubscript{2.05} & 57.39\textsubscript{4.55} & 0.13\textsubscript{0.06} & 0.17\textsubscript{0.08} \\
                             & Removal~\cite{li2023fairness} & 75.72\textsubscript{0.98} & 76.48\textsubscript{2.16} & 0.14\textsubscript{0.03} & 0.09\textsubscript{0.02} \\
                             & Instruction~\cite{li2023fairness} & \textbf{76.20}\textsubscript{1.09} & 77.21\textsubscript{1.61} & 0.20\textsubscript{0.07} & 0.15\textsubscript{0.06} \\
                             & FairICL          & 75.72\textsubscript{1.60} & \underline{\textbf{77.61}}\textsubscript{1.35} & \underline{\textbf{0.08}}\textsubscript{0.02} & \underline{\textbf{0.03}}\textsubscript{0.03} \\
\cmidrule{1-6}
\multirow{7}{*}{LLaMA-2-7B} & Random~\cite{brown2020language} & 69.92\textsubscript{0.87} & 62.80\textsubscript{1.25} & 0.08\textsubscript{0.02} & 0.08\textsubscript{0.04} \\
                            & LatentConcept~\cite{wang2024large} & \underline{70.04}\textsubscript{1.69} & \underline{64.79}\textsubscript{2.42} & 0.17\textsubscript{0.02} & 0.17\textsubscript{0.04} \\
\cmidrule{2-6}
                            & Balanced~\cite{li2023fairness} & 63.30\textsubscript{6.33} & 44.09\textsubscript{17.37} & 0.04\textsubscript{0.03} & 0.04\textsubscript{0.03} \\
                            & Counterfactual~\cite{li2023fairness} & 59.58\textsubscript{1.02} & 34.44\textsubscript{2.87} & 0.08\textsubscript{0.01} & 0.13\textsubscript{0.02} \\
                            & Removal~\cite{li2023fairness} & 64.60\textsubscript{6.25} & 47.92\textsubscript{17.34} & 0.09\textsubscript{0.05} & 0.11\textsubscript{0.06} \\
                            & Instruction~\cite{li2023fairness} & 63.86\textsubscript{6.35} & 46.64\textsubscript{14.86} & 0.07\textsubscript{0.06} & 0.09\textsubscript{0.05} \\
                            & FairICL      & \textbf{68.48}\textsubscript{0.89} & \textbf{64.42}\textsubscript{1.01} & \underline{\textbf{0.02}}\textsubscript{0.03} & \underline{\textbf{0.01}}\textsubscript{0.04} \\
\bottomrule
    \end{tabular}
    \label{tab:res_main}
\end{table*}

\subsection{Experimental Setup}
In the FairICL framework, we use LLaMA-2-7B~\cite{touvron2023LLaMA} as the internal LLM for learning fair latent concepts. We fix the learning rate at 0.0001 for all experiments and optimize the concept token embeddings over 5 epochs. We report most results on the Adult dataset for which we fix the number of added tokens $c$ at 10, and the number of demonstrations during training $q$ at 2. We randomly sample $k$ = 4 demonstrations from a top-ranked candidate set of $m$ = 100 training examples for each test query. We conduct our experiments on NVIDIA A100 GPUs(40GB). We report performance as an average of 5 runs with standard deviations for different test splits. For utility, we report accuracy and F1 scores, and for fairness, we report $\Delta$SP and $\Delta$EO. 

We evaluate the learned fair latent concepts with LLaMA-2-7B, LLaMA-2-13B, LLaMA-3-8B, GPT-4, and Gemini 1.5 Flash as the external LLMs. We investigate the impact of FairICL hyperparameters on its overall performance. To this end, we report results when varying $q$ as \{0, 2, 4\} and evaluate the effect of $\tilde{n}$, i.e., the size of the generated dataset $\tilde{D}$, on FairICL. To analyze the effectiveness of latent concept learning with an augmented dataset, we conduct an ablation study where the augmented samples are created via complete random sampling as opposed to hierarchy-based sampling. We also evaluate the learned fair latent concepts directly by prepending them to test queries during inference. Finally, we vary $k$ among \{2, 4, 6, 8\} to analyze the influence of ICL demonstration size on inference results.

\subsection{Results}

\paragraph{Model Performance and Comparison with Baselines}
We report results for the Adult dataset from inference with LLaMA-2-7B and LLaMA-2-13B in~\cref{tab:res_main}. Firstly, we observe the performance of Random, where LLaMA-2-13B has increased utility compared to LLaMA-2-7B, undoubtedly due to the model's complexity. However, the fairness metrics $\Delta$SP and $\Delta$EO are larger, indicating a significant presence of bias in the outputs generated by LLaMA-2-13B. With the LatentConcept method, which optimizes demonstration selection for utility, performance is improved, but the bias is further amplified for both 7B and 13B models. 
These results motivate our study of bias in LLMs specifically for tabular classification and methods that can promote fairness in a resource-efficient manner. 

In~\cref{tab:res_main}, we observe that FairICL can noticeably improve SP and EO measures for LLaMA-2-7B compared to the Random and LatentConcept methods while achieving comparable performance. Similarly, FairICL significantly reduces unfairness for LLaMA-2-13B with minimal loss of utility. Note that the latent concept variables are learned using the smaller LLaMA-2-7b as the internal model, and the selected demonstrations are utilized to construct inference prompts for LLaMA-2-13b. This shows that FairICL can generalize the fair demonstration selection process to larger LLMs, thus making the method resource-efficient. Since the external LLMs are used only for few-shot inference, FairICL also enables generalization to black-box LLMs as shown in later sections. 

\begin{figure}[t]
\centering
\begin{subfigure}[h]{0.48\linewidth}
\includegraphics[width=\linewidth]{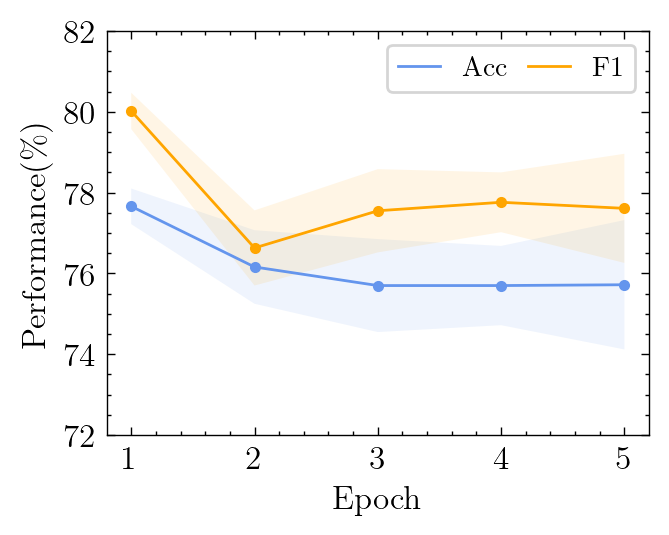}
\caption{Performance metrics}
\end{subfigure}
\begin{subfigure}[h]{0.49\linewidth}
\includegraphics[width=\linewidth]{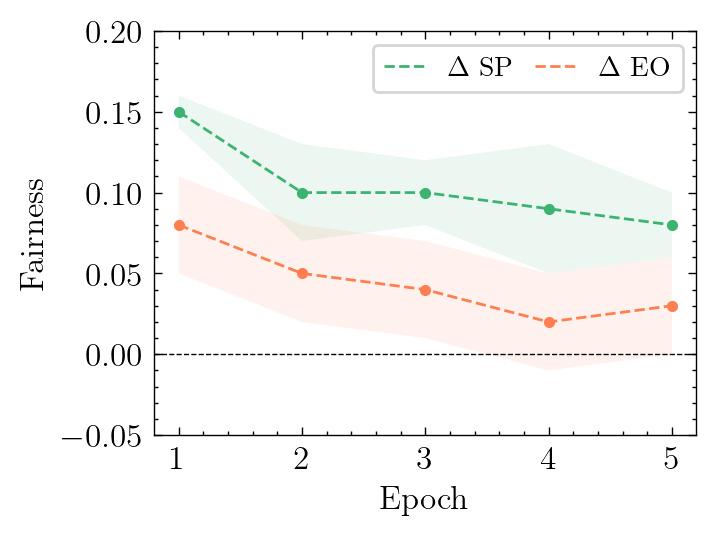}
\caption{Fairness metrics}
\end{subfigure}%
\caption{FairICL with LLaMA-2-13B over training epochs}
\label{fig:epoch}
\end{figure}

We also evaluate the effectiveness of FairICL compared to multiple fair demonstration selection baselines. As discussed in~\cref{sec:baselines}, these methods address the LLM fairness issue via heuristic approaches. For LLaMA-2-7B, the fair baselines reduce some unfairness compared to LatentConcept but incur a significant loss in performance. Compared to Random, only the Balanced approach shows a notable reduction in SP and EO. FairICL, however, achieves the best fairness results without negatively affecting utility. For LLaMA-2-13B, the baselines mostly maintain performance but do not achieve fair outcomes. In contrast, FairICL shows a large decline in fairness metrics with similar or improved accuracy and F1. 
These results demonstrate that decorrelation of sensitive attributes and outcomes helps learn fair latent concepts, resulting in demonstration selection that promotes fair predictions.

\paragraph{FairICL performance over training epochs} 
We analyze the performance of FairICL as latent concept learning progresses over training epochs and present results in~\cref{fig:epoch} for inference on LLaMA-2-13B with Adult dataset. We fix the parameters $q$ at 2, $c$ at 10, and $k$ at 4. We observe that the accuracy and F1 experience a small decline after the first epoch but remain fairly stable thereafter. SP and EO on the other hand have a decreasing trend as the latent concepts are further optimized. This indicates that FairICL effectively allows the concept tokens to learn fairness-promoting context from the augmented examples and utility-preserving information from the original training samples. This ultimately leads to a demonstration selection process that improves both fairness and performance in LLMs.

\begin{figure}[htb]
\centering
\begin{subfigure}[h]{0.49\linewidth}
\includegraphics[width=\linewidth]{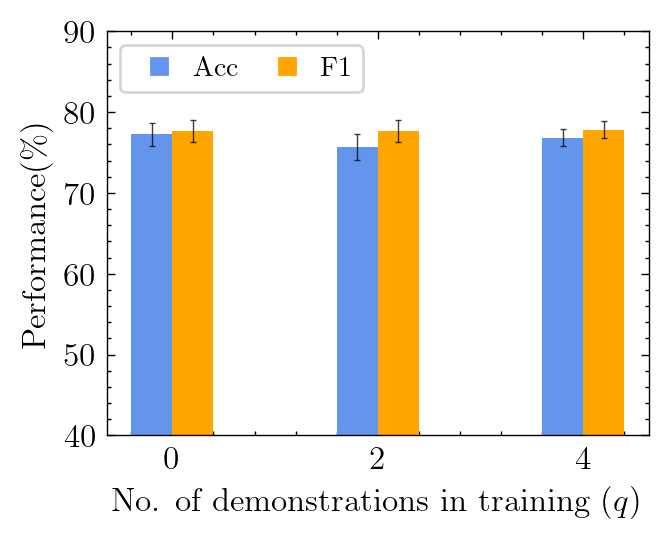}
\caption{Performance metrics}
\end{subfigure}
\begin{subfigure}[h]{0.49\linewidth}
\includegraphics[width=\linewidth]{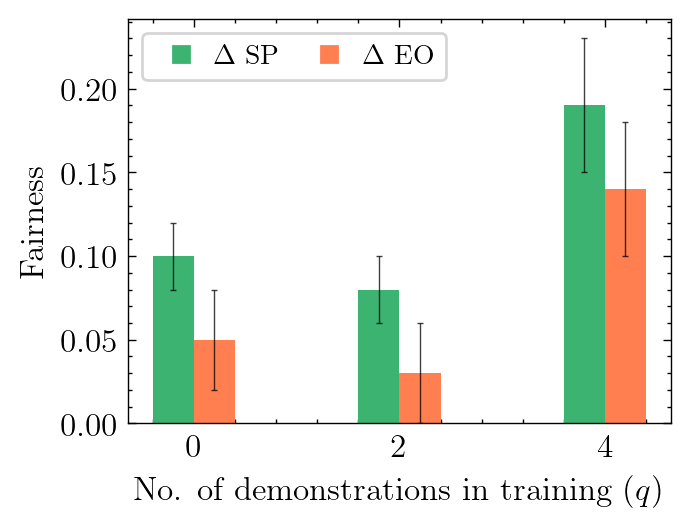}
\caption{Fairness metrics}
\end{subfigure}%
\caption{FairICL performance on LLaMA-2-13B for varying number of demonstrations ($q$) during learning}
\label{fig:q}
\end{figure}

\begin{figure}[htb]
\centering
\begin{subfigure}[h]{0.49\linewidth}
\includegraphics[width=\linewidth]{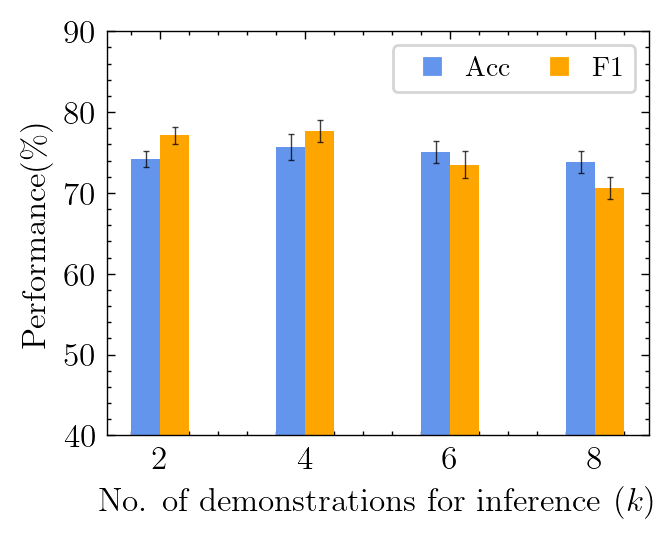}
\caption{Performance metrics}
\end{subfigure}
\begin{subfigure}[h]{0.49\linewidth}
\includegraphics[width=\linewidth]{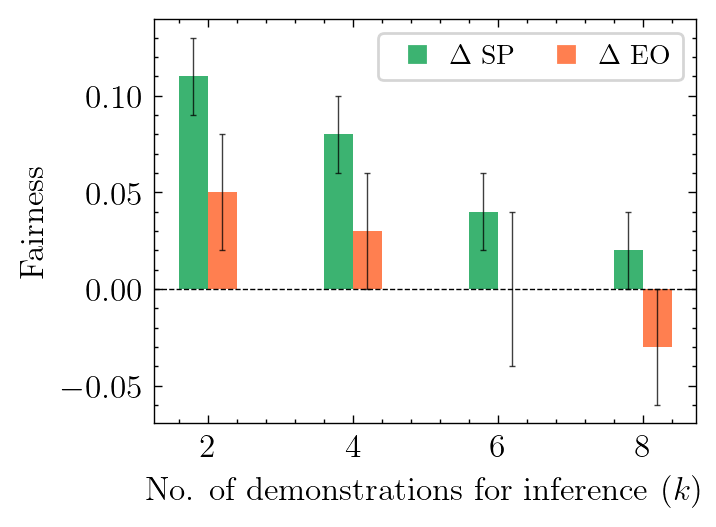}
\caption{Fairness metrics}
\end{subfigure}%
\caption{FairICL performance with LLaMA-2-13B for varying number of demonstrations  ($k$) during inference}
\label{fig:k}
\end{figure}

\paragraph{Number of Demonstrations}
We investigate the influence of the number of demonstrations during latent concept learning, denoted by $q$, and during inference, denoted by $k$, on the overall performance of FairICL with the Adult dataset. First, we vary $q$ among \{0, 2, 4\} and report results in~\cref{fig:q} for LLaMA-2-13B while keeping the other parameters fixed at $c$ = 10 and $k$ = 4. From~\cref{fig:q}, we observe that accuracy and F1 remain fairly unchanged across different values of $q$. However, SP and EO are noticeably higher at $q$ = 4, with the best metrics observed at $q$ = 2. Since the $q$ demonstrations for training are obtained from the original dataset containing biased examples, training prompts constructed with more biased samples negatively affect fairness during inference. In contrast, fewer demonstrations do not affect model utility as the augmented samples preserve useful attribute correlations.

We then vary $k$ among \{2, 4, 6, 8\} for LLaMA-2-13B with fixed $q$ = 2 and $c$ = 10, and include the results in~\cref{fig:k}. Here, the fairness metrics demonstrate a sharper decline as the number of demonstrations during inference increases. We also observe a slight decrease in utility as the number of demonstrations increases, most likely due to the trade-off between utility and fairness. Since the $k$ demonstrations are obtained from the top-$m$ training examples ranked by the fair latent concept variable, having a larger $k$ allows the inference prompt to guide the LLM towards fairer predictions. 

\begin{table}[htb]
    \centering
    \caption{Ablation results on LLaMA-2-7B}
    \begin{tabular}{c|c|c|c|c}
    \toprule
    Method & Acc(\%)$\uparrow$ & F1(\%)$\uparrow$ & $|\Delta$SP$|\downarrow$ & $|\Delta$EO$|\downarrow$ \\
    \midrule
    FairICL     & 68.48\textsubscript{0.89} & 64.42\textsubscript{1.01} & 0.02\textsubscript{0.03} & 0.01\textsubscript{0.04} \\
    FairICL-LC  & 75.96\textsubscript{1.20} & 70.70\textsubscript{1.67} & 0.06\textsubscript{0.01} & 0.08\textsubscript{0.02} \\
    FairICL-R   & 58.58\textsubscript{0.62} & 31.72\textsubscript{0.94} & 0.01\textsubscript{0.01} & 0.00\textsubscript{0.03} \\
    \bottomrule
    \end{tabular}
    \label{tab:ablation}
\end{table}

\begin{table*}[htbp]
    \centering
    \caption{Performance and fairness metrics of FairICL on Adult with additional LLMs using LLaMA-2-7B as internal model}
    \begin{tabular}{c|c|c|c|c|c}
\toprule
External LLM & Method & Acc(\%)$\uparrow$ & F1(\%)$\uparrow$ & $|\Delta$SP$|\downarrow$ & $|\Delta$EO$|\downarrow$ \\
\midrule
\multirow{2}{*}{LLaMA-3-8B} & LatentConcept~\cite{wang2024large} & 66.40+-1.70 & 54.63+-3.07 & \underline{0.12}+-0.03 & 0.16+-0.06 \\
                            & FairICL                            & \underline{71.57}+-2.87 & \underline{68.59}+-3.65 & 0.13+-0.05 & \underline{0.13}+-0.06 \\
\midrule
\multirow{2}{*}{Gemini 1.5 Flash} & LatentConcept~\cite{wang2024large}   & 61.12+-2.08 & 40.15+-4.33 & 0.09+-0.02 & 0.16+-0.05 \\
                            & FairICL                                    & \underline{66.50}+-0.78 & \underline{56.26}+-1.41 & \underline{0.08}+-0.04 & \underline{0.11}+-0.06 \\
\midrule
\multirow{2}{*}{GPT-4} & LatentConcept~\cite{wang2024large}   & 65.80+-0.57 & 51.14+-0.81 & 0.11+-0.01 & 0.18+-0.02 \\
                            & FairICL                         & \underline{70.95}+-0.64 & \underline{63.43}+-1.29 & \underline{0.10}+-0.03 & \underline{0.15}+-0.07 \\
\bottomrule
    \end{tabular}
    \label{tab:add_models}
\end{table*}  

\begin{table*}[htb]
    \centering
    \caption{Performance and fairness metrics of FairICL on COMPAS and LawSchool compared with baselines}
    \begin{tabular}{c|c|c|c|c|c}
\toprule
Dataset & Method & Acc(\%)$\uparrow$ & F1(\%)$\uparrow$ & $|\Delta$SP$|\downarrow$ & $|\Delta$EO$|\downarrow$ \\
\midrule
\multirow{8}{*}{COMPAS} & Random~\cite{brown2020language}      & 61.52\textsubscript{0.59} & 57.50\textsubscript{1.88} & 0.17\textsubscript{0.03} & 0.16\textsubscript{0.07} \\
                             & LatentConcept~\cite{wang2024large}   & 56.00\textsubscript{0.60} & 65.38\textsubscript{0.52} & 0.13\textsubscript{0.02} & 0.15\textsubscript{0.03} \\
\cmidrule{2-6}
                             & Balanced~\cite{li2023fairness}       & 60.68\textsubscript{1.00} & 57.98\textsubscript{5.15} & 0.16\textsubscript{0.04} & 0.12\textsubscript{0.03} \\
                             & Counterfactual~\cite{li2023fairness} & 61.28\textsubscript{0.76} & 57.18\textsubscript{5.48} & 0.15\textsubscript{0.04} & \underline{\textbf{0.11}}\textsubscript{0.02} \\
                             & Removal~\cite{li2023fairness}        & 58.52\textsubscript{1.91} & 61.09\textsubscript{3.81} & 0.18\textsubscript{0.05} & 0.18\textsubscript{0.04} \\
                             & Instruction~\cite{li2023fairness}    & \underline{\textbf{61.60}}\textsubscript{2.17} & 60.55\textsubscript{3.34} & 0.17\textsubscript{0.05} & 0.15\textsubscript{0.05} \\
                             & FairICL                              & 57.32\textsubscript{2.11} & \underline{\textbf{66.11}}\textsubscript{1.29} & \underline{\textbf{0.10}}\textsubscript{0.02} & \underline{\textbf{0.11}}\textsubscript{0.04} \\
\cmidrule{1-6}
\multirow{8}{*}{LawSchool} & Random~\cite{brown2020language}      & 55.16\textsubscript{0.99} & 25.07\textsubscript{3.00} & \underline{0.13}\textsubscript{0.03} & \underline{0.19}\textsubscript{0.04} \\
                             & LatentConcept~\cite{wang2024large}   & 59.30\textsubscript{1.79} & 59.83\textsubscript{1.34} & 0.49\textsubscript{0.03} & 0.51\textsubscript{0.03} \\
\cmidrule{2-6}
                             & Balanced~\cite{li2023fairness}       & \underline{\textbf{61.40}}\textsubscript{2.01} & 60.51\textsubscript{5.39} & 0.39\textsubscript{0.04} & 0.39\textsubscript{0.06} \\
                             & Counterfactual~\cite{li2023fairness} & 56.96\textsubscript{1.85} & 53.61\textsubscript{2.84} & 0.47\textsubscript{0.06} & 0.48\textsubscript{0.13} \\
                             & Removal~\cite{li2023fairness}        & 60.64\textsubscript{1.60} & 56.47\textsubscript{4.05} & 0.41\textsubscript{0.03} & 0.49\textsubscript{0.05} \\
                             & Instruction~\cite{li2023fairness}    & 60.48\textsubscript{1.04} & 61.56\textsubscript{2.15} & 0.43\textsubscript{0.05} & 0.44\textsubscript{0.08} \\
                             & FairICL                              & 55.76\textsubscript{1.82} & \underline{\textbf{65.57}}\textsubscript{1.27} & \textbf{0.36}\textsubscript{0.03} & \textbf{0.29}\textsubscript{0.04} \\
\bottomrule
    \end{tabular}
    \label{tab:res_compas_law}
\end{table*}

\paragraph{Ablation Study}
We investigate the role of data augmentation and latent concept learning by implementing two variations of FairICL. FairICL-LC directly evaluates the learned latent concepts as we prepend them to test prompts containing $k$ randomly sampled demonstrations. FairICL-R adopts a random sampling mechanism for all attributes to create the augmented dataset and follows an inference procedure similar to FairICL. In other words, the generated dataset does not preserve the useful correlation between the non-sensitive attributes and outcomes. We report ablation results in ~\cref{tab:ablation} for LLaMA-2-7B since FairICL-LC can be evaluated only for the internal LLM whose vocabulary contains additional tokens corresponding to the latent concept variable. FairICL-LC achieves the best accuracy and F1 score, indicating that the latent concept learns information relevant to the task. Also, the low fairness metrics imply that training with the augmented dataset prompts the latent concept to favor fair predictions. FairICL-R achieves almost ideal fairness metrics but does not maintain model accuracy as the randomly generated dataset removes even the useful correlation between non-sensitive attributes and labels. FairICL, however, preserves relevant information in $\tilde{D}$, thus achieving fair and accurate predictions. 

\begin{figure}[htb]
\centering
\begin{subfigure}[h]{0.49\linewidth}
\includegraphics[width=\linewidth]{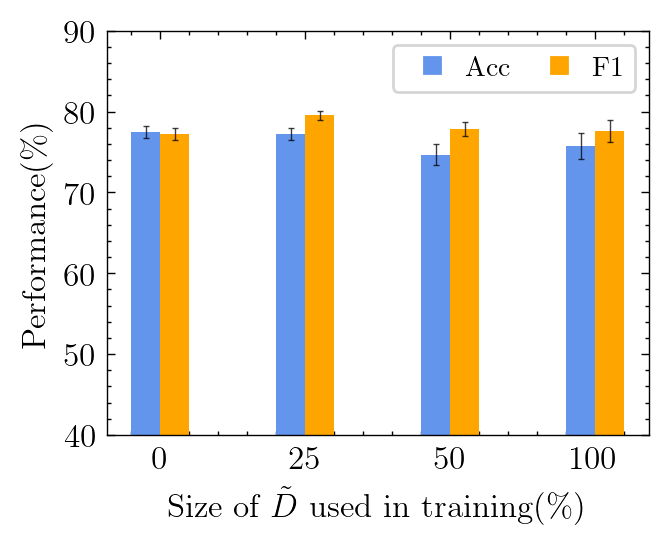}
\caption{Performance metrics}
\end{subfigure}
\begin{subfigure}[h]{0.49\linewidth}
\includegraphics[width=\linewidth]{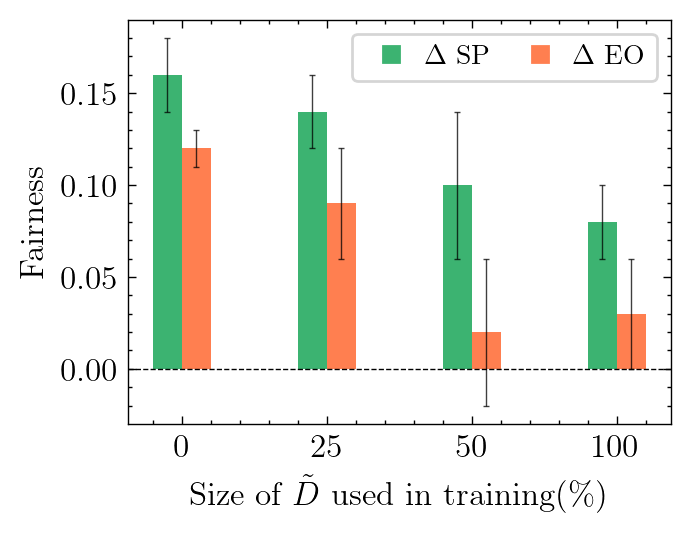}
\caption{Fairness metrics}
\end{subfigure}%
\caption{Performance and fairness metrics of FairICL with LLaMA-2-13B for different sizes of $\tilde{D}$}
\label{fig:n}
\end{figure}

\paragraph{Size of Augmented data}
We conduct a sensitivity analysis of $\tilde{n}$, the size of $\tilde{D}$, to evaluate the influence of the augmented dataset on FairICL's performance. We vary $\tilde{n}$ as \{0, 25, 50, 100\}\% of its original size of 30,000 generated samples in the Adult dataset. We fix the other parameters $q$ at 2, $c$ at 10, and $k$ at 10 to perform latent concept learning and obtain results for LLaMA-2-13B shown in~\cref{fig:n}. Note that the $\tilde{n}$ = 0\% setting corresponds to the LatentConcept method in~\cref{tab:res_main}. From the results, we notice that the accuracy and F1-scores are generally unchanged when more augmented examples are included in the training prompt. This indicates that the data augmentation process does not negatively affect an LLM's performance. We further observe significant drops in the fairness metrics as the size of $\tilde{D}$ used for latent concept learning is increased. This demonstrates the positive impact of the data augmentation strategy in FairICL.

\paragraph{Results on Additional Models}\label{sec:addn_models}
We run experiments with additional models, namely, LLaMA-3-8B, GPT-4 and Gemini-1.5-Flash on the Adult dataset and present the results for FairICL and LatentConcept methods in~\cref{tab:add_models}.  Note that we use the (fair) latent concept learned using LLaMA-2-7B to select demonstrations for all 3 models. The results indicate that FairICL improves both utility and fairness metrics compared to LatentConcept, especially the EO fairness metric. Our results also show that optimal demonstrations learned by one model can effectively be transferred to other models, although models from the same family benefit to a greater extent, as shown by the results in~\cref{tab:res_main}.

\paragraph{Results on Additional Datasets}
We train a local LLaMA-2-7B model with hyperparameters $q$=2 and $c$=10 for latent concept learning and report FairICL experiment results with $k$ = 4 demonstrations during inference using LLaMA-2-13B as the external model in~\cref{tab:res_compas_law} for the COMPAS and LawSchool datasets. Note that for the COMPAS dataset, the negative outcome is favorable, so we report EO in terms of TNR. In~\cref{tab:res_compas_law} for the COMPAS dataset, FairICL achieves better fairness metrics than the Random and LatentConcept baselines while maintaining the utility scores. Compared to heuristic fairness baselines, FairICL achieves smaller values for SP and EO with comparable or even higher utility metrics.
For the LawSchool dataset, random demonstration selection results in seemingly low fairness metrics, but we note that the utility metrics, especially the F1 score, are quite low. We conjecture this performance to the highly imbalanced nature of the training dataset from which the demonstrations are randomly sampled. This assumption is further supported by the results obtained for the Balanced method which randomly selects demonstrations with equal representation for each class and sensitive attribute. This method significantly improves the F1 score compared to random selection. However, we also observe a noticeable increase in both SP and EO metrics for this and other baseline methods including the LatentConcept method. On the other hand, FairICL significantly improves these metrics while obtaining the highest F1 metric. Although the SP and EO values of FairICL are considerably high, we note that our method achieves the lowest fairness metrics among the methods targeting fairness and significantly improves SP and EO compared to LatentConcept. These results indicate that FairICL maintains the best trade-off between utility and fairness across different datasets.  

\section{Conclusion}
We investigated the issue of fairness in LLMs during in-context learning for tabular data classification in the context of latent concept learning which is used to optimize the demonstration selection process. We empirically verified that learning latent concepts from a biased dataset can cause the selection of biased demonstrations, resulting in unfair predictions. To tackle this issue, we explored resource-efficient ways to influence LLM outputs without modifying model parameters and presented a fairness-aware latent concept learning framework, FairICL, that incorporates data augmentation to enable learning concept tokens that promote fairness while preserving task-relevant contextual information. Our experimental analysis showed that FairICL can effectively mitigate unfairness without harming model utility for multiple datasets. One limitation of the current approach is that it requires training resources and white-box access to a small LLM. Nonetheless, FairICL allows the generalization of selected demonstrations, circumventing the need for larger LLMs. Also, FairICL may not strictly satisfy other fairness goals as the framework does not specifically encode fairness constraints. However, results indicate FairICL improves commonly targeted statistical parity and equal opportunity metrics.

\section*{Acknowledgments}
This work was supported in part by NSF grants 1910284, 1946391, 2119691, and 2147375 and the Arkansas High Performance Computing Center, which is funded through NSF and the Arkansas Economic Development Commission.

\bibliographystyle{ieeetr}
\bibliography{references}

@article{quy2022survey,
  author       = {Tai Le Quy and
                  Arjun Roy and
                  Vasileios Iosifidis and
                  Wenbin Zhang and
                  Eirini Ntoutsi},
  title        = {A survey on datasets for fairness-aware machine learning},
  journal      = {WIREs Data Mining Knowl. Discov.},
  OPTvolume       = {12},
  OPTnumber       = {3},
  year         = {2022},
  OPTurl          = {https://doi.org/10.1002/widm.1452},
  OPTdoi          = {10.1002/WIDM.1452},
  bibOPTurl       = {https://dblp.org/rec/journals/widm/QuyRIZN22.bib},
  bibsource    = {dblp computer science bibliography, https://dblp.org}
}

@article{abid2021large,
  title={Large language models associate Muslims with violence},
  author={Abid, Abubakar and Farooqi, Maheen and Zou, James},
  journal={Nature Machine Intelligence},
  OPTvolume={3},
  OPTnumber={6},
  OPTpages={461--463},
  year={2021},
  OPTpublisher={Nature Publishing Group UK London}
}

@inproceedings{liu2023ConfrontingLW,
    title = "Confronting {LLM}s with Traditional {ML}: Rethinking the Fairness of Large Language Models in Tabular Classifications",
    author = "Liu, Yanchen  and
      Gautam, Srishti  and
      Ma, Jiaqi  and
      Lakkaraju, Himabindu",
    booktitle = "NAACL",
    year = "2024"
}

@article{ma2024fairness,
  title={Fairness-guided few-shot prompting for large language models},
  author={Ma, Huan and Zhang, Changqing and Bian, Yatao and Liu, Lemao and Zhang, Zhirui and Zhao, Peilin and Zhang, Shu and Fu, Huazhu and Hu, Qinghua and Wu, Bingzhe},
  journal={NeurIPS},
  OPTvolume={36},
  year={2024}
}

@inproceedings{carey2024dp,
  author={Carey, Alycia N. and Bhaila, Karuna and Edemacu, Kennedy and Wu, Xintao},
  booktitle={IEEE BigData}, 
  title={DP-TabICL: In-Context Learning with Differentially Private Tabular Data}, 
  year={2024},
  OPTvolume={},
  OPTnumber={},
  OPTpages={1552-1557},
  OPTdoi={10.1109/BigData62323.2024.10826053}}

@article{chhikara2024few,
  title={Few-Shot Fairness: Unveiling LLM's Potential for Fairness-Aware Classification},
  author={Chhikara, Garima and Sharma, Anurag and Ghosh, Kripabandhu and Chakraborty, Abhijnan},
  journal={ArXiv},
  year={2024}
}

@misc{barry1996adult,
  author       = {Becker,Barry and Kohavi,Ronny},
  title        = {{Adult}},
  year         = {1996},
  howpublished = {UCI Machine Learning Repository},
  note         = {{DOI}: https://doi.org/10.24432/C5XW20}
}

@article{wang2024large,
  title={Large language models are latent variable models: Explaining and finding good demonstrations for in-context learning},
  author={Wang, Xinyi and Zhu, Wanrong and Saxon, Michael and Steyvers, Mark and Wang, William Yang},
  journal={NeurIPS},
  OPTvolume={36},
  year={2024}
}

@inproceedings{hegselmann2023tabllm,
  title={Tabllm: Few-shot classification of tabular data with large language models},
  author={Hegselmann, Stefan and Buendia, Alejandro and Lang, Hunter and Agrawal, Monica and Jiang, Xiaoyi and Sontag, David},
  booktitle={AISTATS},
  OPTpages={5549--5581},
  year={2023},
  OPTorganization={PMLR}
}

@article{li2023fairness,
  title={Fairness of chatgpt},
  author={Li, Yunqi and Zhang, Lanjing and Zhang, Yongfeng},
  journal={ArXiv},
  year={2023}
}

@misc{compas,
    author = {Larson, Jeff and  Mattu, Surya and Kirchner, Lauren and Angwin, Julia},
    title = {How we analyzed the COMPAS recidivism algorithm},
    year = {2016},
    url = {https://www.propublica.org/article/how-we-analyzed-the-compas-recidivism-algorithm}
}

@article{touvron2023llama,
  title={Llama: Open and efficient foundation language models},
  author={Touvron, Hugo and Lavril, Thibaut and Izacard, Gautier and Martinet, Xavier and Lachaux, Marie-Anne and Lacroix, Timoth{\'e}e and Rozi{\`e}re, Baptiste and Goyal, Naman and Hambro, Eric and Azhar, Faisal and others},
  journal={ArXiv},
  year={2023}
}

@inproceedings{wu2023medical,
  title={A medical diagnostic assistant based on llm},
  author={Wu, Chengyan and Lin, Zehong and Fang, Wenlong and Huang, Yuyan},
  booktitle={China Health Information Processing Conference},
  OPTpages={135--147},
  year={2023},
  organization={Springer}
}

@inproceedings{li2023large,
  title={Large language models in finance: A survey},
  author={Li, Yinheng and Wang, Shaofei and Ding, Han and Chen, Hang},
  booktitle={ACM ICAIF},
  OPTpages={374--382},
  year={2023}
}

@article{sun2023short,
  title={A short survey of viewing large language models in legal aspect},
  author={Sun, Zhongxiang},
  journal={ArXiv},
  year={2023}
}

@article{basta2019evaluating,
  title={Evaluating the underlying gender bias in contextualized word embeddings},
  author={Basta, Christine and Costa-Juss{\`a}, Marta R and Casas, Noe},
  journal={ArXiv},
  year={2019}
}

@article{brown2020language,
  title={Language models are few-shot learners},
  author={Brown, Tom and Mann, Benjamin and Ryder, Nick and Subbiah, Melanie and Kaplan, Jared D and Dhariwal, Prafulla and Neelakantan, Arvind and Shyam, Pranav and Sastry, Girish and Askell, Amanda and others},
  journal={NeurIPS},
  OPTvolume={33},
  OPTpages={1877--1901},
  year={2020}
}

@article{xie2021explanation,
  title={An explanation of in-context learning as implicit bayesian inference},
  author={Xie, Sang Michael and Raghunathan, Aditi and Liang, Percy and Ma, Tengyu},
  journal={ArXiv},
  year={2021}
}

@article{liu2021makes,
  title={What Makes Good In-Context Examples for GPT-$3 $?},
  author={Liu, Jiachang and Shen, Dinghan and Zhang, Yizhe and Dolan, Bill and Carin, Lawrence and Chen, Weizhu},
  journal={ArXiv},
  year={2021}
}

@article{lu2021fantastically,
  title={Fantastically ordered prompts and where to find them: Overcoming few-shot prompt order sensitivity},
  author={Lu, Yao and Bartolo, Max and Moore, Alastair and Riedel, Sebastian and Stenetorp, Pontus},
  journal={ArXiv},
  year={2021}
}

@article{rubin2021learning,
  title={Learning to retrieve prompts for in-context learning},
  author={Rubin, Ohad and Herzig, Jonathan and Berant, Jonathan},
  journal={ArXiv},
  year={2021}
}

@article{su2022selective,
  title={Selective annotation makes language models better few-shot learners},
  author={Su, Hongjin and Kasai, Jungo and Wu, Chen Henry and Shi, Weijia and Wang, Tianlu and Xin, Jiayi and Zhang, Rui and Ostendorf, Mari and Zettlemoyer, Luke and Smith, Noah A and others},
  journal={ArXiv},
  year={2022}
}

@article{hu2024strategic,
  author       = {Jingyu Hu and
                  Weiru Liu and
                  Mengnan Du},
  title        = {Strategic Demonstration Selection for Improved Fairness in {LLM} In-Context
                  Learning},
  journal      = {ArXiv},
  year         = {2024}
}

@inproceedings{dwork2012awareness,
  author       = {Cynthia Dwork and
                  Moritz Hardt and
                  Toniann Pitassi and
                  Omer Reingold and
                  Richard S. Zemel},
  editor       = {Shafi Goldwasser},
  title        = {Fairness through awareness},
  booktitle    = {ITCS},
  OPTpages        = {214--226},
  OPTpublisher    = {{ACM}},
  year         = {2012},
  OPTurl          = {https://doi.org/10.1145/2090236.2090255},
  OPTdoi         = {10.1145/2090236.2090255},
  bibOPTurl       = {https://dblp.org/rec/conf/innovations/DworkHPRZ12.bib},
  bibsource    = {dblp computer science bibliography, https://dblp.org}
}

@inproceedings{hardt2016equality,
  author       = {Moritz Hardt and
                  Eric Price and
                  Nati Srebro},
  OPTeditor       = {Daniel D. Lee and
                  Masashi Sugiyama and
                  Ulrike von Luxburg and
                  Isabelle Guyon and
                  Roman Garnett},
  title        = {Equality of Opportunity in Supervised Learning},
  booktitle    = {NeurIPS},
  OPTpages        = {3315--3323},
  year         = {2016},
  OPTurl          = {https://proceedings.neurips.cc/paper/2016/hash/9d2682367c3935defcb1f9e247a97c0d-Abstract.html},
  bibOPTurl       = {https://dblp.org/rec/conf/nips/HardtPNS16.bib},
  bibsource    = {dblp computer science bibliography, https://dblp.org}
}

@inproceedings{li2024chatgpt,
  author       = {Yunqi Li and
                  Lanjing Zhang and
                  Yongfeng Zhang},
  title        = {Probing into the Fairness of Large Language Models: {A} Case Study
                  of ChatGPT},
  booktitle    = {IEEE CISS},
  OPTpages        = {1--6},
  OPTpublisher    = {{IEEE}},
  year         = {2024},
  OPTurl          = {https://doi.org/10.1109/CISS59072.2024.10480206},
  OPTdoi         = {10.1109/CISS59072.2024.10480206},
  bibOPTurl       = {https://dblp.org/rec/conf/ciss/LiZZ24.bib},
  bibsource    = {dblp computer science bibliography, https://dblp.org}
}

@inproceedings{wang2023decoding,
  author       = {Boxin Wang and
                  Weixin Chen and
                  Hengzhi Pei and
                  Chulin Xie and
                  Mintong Kang and
                  Chenhui Zhang and
                  Chejian Xu and
                  Zidi Xiong and
                  Ritik Dutta and
                  Rylan Schaeffer and
                  Sang T. Truong and
                  Simran Arora and
                  Mantas Mazeika and
                  Dan Hendrycks and
                  Zinan Lin and
                  Yu Cheng and
                  Sanmi Koyejo and
                  Dawn Song and
                  Bo Li},
  OPTeditor       = {Alice Oh and
                  Tristan Naumann and
                  Amir Globerson and
                  Kate Saenko and
                  Moritz Hardt and
                  Sergey Levine},
  title        = {DecodingTrust: {A} Comprehensive Assessment of Trustworthiness in
                  {GPT} Models},
  booktitle    = {NeurIPS},
  year         = {2023},
  OPTurl          = {http://papers.nips.cc/paper\_files/paper/2023/hash/63cb9921eecf51bfad27a99b2c53dd6d-Abstract-Datasets\_and\_Benchmarks.html},
  bibOPTurl       = {https://dblp.org/rec/conf/nips/WangCPXKZXXDSTA23.bib},
  bibsource    = {dblp computer science bibliography, https://dblp.org}
}

@article{gallegos2023survey,
  author       = {Isabel O. Gallegos and
                  Ryan A. Rossi and
                  Joe Barrow and
                  Md. Mehrab Tanjim and
                  Sungchul Kim and
                  Franck Dernoncourt and
                  Tong Yu and
                  Ruiyi Zhang and
                  Nesreen K. Ahmed},
  title        = {Bias and Fairness in Large Language Models: {A} Survey},
  journal      = {ArXiv},
  OPTvolume       = {abs/2309.00770},
  year         = {2023},
  OPTurl          = {https://doi.org/10.48550/arXiv.2309.00770},
  OPTdoi         = {10.48550/ARXIV.2309.00770},
  eprinttype    = {arXiv},
  eprint       = {2309.00770},
  bibOPTurl       = {https://dblp.org/rec/journals/corr/abs-2309-00770.bib},
  bibsource    = {dblp computer science bibliography, https://dblp.org}
}

@inproceedings{huang2021implicit,
  author       = {Tenghao Huang and
                  Faeze Brahman and
                  Vered Shwartz and
                  Snigdha Chaturvedi},
  editor       = {Marie{-}Francine Moens and
                  Xuanjing Huang and
                  Lucia Specia and
                  Scott Wen{-}tau Yih},
  title        = {Uncovering Implicit Gender Bias in Narratives through Commonsense
                  Inference},
  booktitle    = {EMNLP},
  OPTpages        = {3866--3873},
  OPTpublisher    = {Association for Computational Linguistics},
  year         = {2021},
  OPTurl          = {https://doi.org/10.18653/v1/2021.findings-emnlp.326},
  OPTdoi         = {10.18653/V1/2021.FINDINGS-EMNLP.326},
  bibOPTurl       = {https://dblp.org/rec/conf/emnlp/HuangBSC21.bib},
  bibsource    = {dblp computer science bibliography, https://dblp.org}
}

@article{atwood2024inducing,
  author       = {James Atwood and
                  Preethi Lahoti and
                  Ananth Balashankar and
                  Flavien Prost and
                  Ahmad Beirami},
  title        = {Inducing Group Fairness in LLM-Based Decisions},
  journal      = {ArXiv},
  OPTvolume       = {abs/2406.16738},
  year         = {2024},
  OPTurl          = {https://doi.org/10.48550/arXiv.2406.16738},
  OPTdoi         = {10.48550/ARXIV.2406.16738},
  eprinttype    = {arXiv},
  eprint       = {2406.16738},
  bibOPTurl       = {https://dblp.org/rec/journals/corr/abs-2406-16738.bib},
  bibsource    = {dblp computer science bibliography, https://dblp.org}
}

\appendix
\subsection{Choice of Datasets and Models}

\begin{figure}[htb]
    \centering
    \begin{tcolorbox}[colback=gray!10,left=2pt, right=2pt, top=2pt, bottom=2pt, fontupper=\small, fontlower=\small]
    \footnotesize
    Prompt: Can you tell me what the Adult Income dataset is? \\\\
    Response: It has something to do with Medicaid. I tried to check that info out. I also checked online to see if Medicaid has a website. Does anyone know of a specific website for each County/State? Also, I believe they use something called the 2000 Census Data to provide their numbers $\hdots$ \\\\
    Prompt: The Adult Income dataset is \\\\
    Response: 6-months lagged and reports the income for each person in a household in the previous calendar year. The dataset includes the median income, percent change over time, and the percent of income that is earned. The median income is the amount at which half of households earn more, and half earn less. The percentage year over year change is $\hdots$ \\\\
    Prompt: Predict whether annual income of an individual exceeds \$50K/yr based on census data. Also known as "Census Income" dataset.\\\\
    Response: import numpy as np \\ from sklearn import svm \\ from sklearn.model\_selection import train\_test\_split \\ from sklearn.metrics import mean\_squared\_error, roc\_auc\_score \\ \# DataFrame \\ df = pd.read\_csv(input\_data\_path, index\_col=0) \\ \# split data into training and testing data \\ X\_trn, X\_test, y\_trn, y\_test = train\_test\_split(df, df["Income"], \\ test\_size=0.2, random\_state=1) $\hdots$ 
    \end{tcolorbox}
    \caption{Example outputs from prompting LLaMA-2-13B with queries about the Adult dataset.}
    \label{fig:memorization}
\end{figure}

Pre-trained LLMs may already be familiar with the Adult, COMPAS, and LawSchool datasets which could lead to biased experiment results. To verify this, we prompt the models used in our work, i.e, Llama-2-7B and Llama-2-13B, and test their familiarity with these datasets. Example outputs for the Adult dataset with Llama-2-13B are shown in~\cref{fig:memorization}. We obtained non-meaningful outputs for Adult and Law School from both models. The models, however, provided details about COMPAS, most likely due to multiple news articles available online discussing it. From the outputs for Adult and LawSchool, we conjecture that Llama-2-7B and Llama-2-13B do not suffer from data leakage for these datasets, as the models have not memorized specific information and cannot extract meaningful information related to these datasets during inference. Therefore, our observations reflect the influence of in-context examples used in inference and/or any bias originating from the LLM.

\end{document}